# Image-Based Path-Planning for Navigation of Soft-Growing Robots in the Spinal Sub-Arachnoid Space*

George Dimas, Panagiotis G. Kalozoumis, Panagiotis Vartholomeos, and Dimitris K. Iakovidis, *Senior Member, IEEE*

*Abstract*— **Recently, soft robotic procedures are being developed aiming at achieving targeted minimally invasive operations or drug administration in specific locations in the human body. Accordingly, specialized path-planning algorithms need to be investigated for the estimation of efficient paths that would protect the delicate tissue structures and simultaneously satisfy a set of requirements for the efficient navigation of the robot. In this context, we propose an image-based 3D path-planning algorithm, which is an extension of the A* algorithm, for the navigation of a soft-growing robot inside the spinal sub-arachnoid space (SSS). The proposed algorithm is capable of estimating a safe pathway towards the goal location, while ensuring the establishment of anchor points that facilitate the efficient steering of the soft robot. The algorithm is evaluated using a highly detailed model of the SSS with respect to its capacity to satisfy the robot movement requirements and path efficiency.**

## I. Introduction

Current research on minimally invasive operations has focused on the development of miniaturized soft robotic systems, aiming at delivering drugs, cells or viruses at hard-to-reach anatomical sites [1]. In this context, the SoftReach[1] project develops a soft growing robot that will provide clinically innovative access to deep brain regions via the spinal sub-arachnoid space (SSS). Specifically, a miniature soft steerable robot, driven by biomimetic apical extension methods, will safely navigate through the SSS under Magnetic Resonance Imaging (MRI) guidance and reach target locations in brain ventricles. There, a drug-loaded scaffold will be attached to the brain ventricle walls. Once attached, the scaffold will release its therapeutic content (*e.g.* pharmaceuticals, cells) into adjacent brain regions. This interventional approach will offer higher therapeutic efficacy and reduced side-effects. Localized delivery of therapeutics to specific deep brain regions can revolutionize the treatment of many neurological disorders that current therapies cannot adequately address [2].

SSS is a complex region which encompasses various anatomical structures, including the spinal cord, dorsal and ventral nerve rootlets, and blood vessels. It is filled with cerebrospinal fluid (CSF), which plays an essential role in protecting the spinal cord by absorbing shocks and preventing the delicate neural tissues from being mechanically damaged. Hence, to enable the robot to reach the target location inside this complex environment, a precise path needs to be generated. Such a path needs to simultaneously ensure that the delicate tissue structures of the SSS are protected and satisfy a set of requirements for the successful navigation of the robot to its target.

Several path-planning algorithms have been proposed to estimate navigation paths for robotic medical instruments such as robotic catheters, steerable needles, and endoscopes [3]. Path-planning is vital to robotic systems used in medical operations, especially when robotic instruments navigate through complex environments comprising sensitive tissue structures. In such cases, path-planning approaches should incorporate strict safety measures.

In general, path-planning methods are categorized into four subgroups: node-based, sampling-based, optimization-based, and learning-based techniques. Methods operating on a node basis, *e.g.*, A-star (A*) and Lifelong Planning A*, have been proposed for path-planning applications for nasal endoscopy, voxel-based shortest-airway-path identification in bronchus models, and insertion pose optimization in brain surgery [4–7]. Similar approaches have been employed for path-planning in intraluminal and endovascular environments using the Dijkstra and Potential field methods [8], [9]. In the context of catheter navigation planning, Fagogenis *et al*. employed a wall following algorithm, efficient for environments where the target can be reached via a limited number of pathways [10].

In the context of path-planning of steerable robots and medical instruments inside the human body, sampling-based methods usually adopt the Randomly Exploring Tree (RTT) algorithm and its variants [11], [12]. Such methodologies have been used in biomedical applications for robotic instruments that follow curvature-constrained trajectories [13]. Optimization- and learning-based methodologies for path planning in medical procedures employing autonomous or collaborative robots have also been proposed. Recently, these methodologies have incorporated approaches such as imitation learning and generative adversarial networks for the navigation of catheters [14]. Usually, path-planning approaches aim at avoiding obstacles while moving towards the target location; thus, they have been mostly applied to environments where the path extends along intraluminal or endovascular surfaces. The presence of obstacles is limited in such environments, and the main objective is to create a path that follows the desired route while avoiding contact with the surrounding tissues that may lead to tissue damage.

* This work has been funded by the European Union, under grant agreement No 101099145, project SoftReach.

G. Dimas, P. G. Kalozoumis, P. Vartholomeos and D. K. Iakovidis are with the Dept. of Computer Science and Biomedical Informatics, School of Science, University of Thessaly, Lamia, Greece (email: {gdimas, pkalozoumis, pvartholomeos, diakovidis}@uth.gr).

[1]https://softreach.eu

In this work, the soft-growing robot consists of two main parts, a steerable tendon-driven catheter and a soft growing element (GE) surrounding the catheter (Fig. 1). Contrary to the aforementioned approaches, only the catheter is designed to avoid collision with critical anatomical structures, while the GE needs to establish anchor points via contact with the surrounding structures. Anchor points are defined as the locations where two opposite sides of the soft GE are in contact with the walls or obstacles (nerves or vasculature) of the SSS. These anchor points are needed to ensure that the maneuvering precision of the catheter is maintained at high levels along the entire estimated path. Although the robust navigation of robots through contact with surrounding structures that are perceived as obstacles has already been investigated [15], the navigation space was simplified into a 2D projection. Nevertheless, the SSS has a complex 3D structure that cannot be simplified into a 2D projection. To this end, this work considers MRI scans as the main localization method for the navigation of the soft robot. The stack of images forms a volume, enabling the estimation of the navigation path in the actual 3D domain.

To cope with the described navigation issues in SSS, this paper investigates an extended version of A* that considers the limitations imposed to the soft-growing robot regarding its maneuverability, *i.e.*, the requirement of establishing anchor points at various locations. The A* algorithm was selected as the basis for the path-planning estimation task, since it is commonly used in static environments, *e.g.*, MRI volumes, and it can converge fast while ensuring optimality [3]. Moreover, A* applications have shown better performance to other path-planning approaches [16], [17]. A twofold extension of A* is investigated: a) a distance-based cost function awarding nodes near obstacles, and b) a parametric cost function concerning the existence of possible anchor points that occur at nodes under investigation. Via these additional cost functions, the novel A* extension is tasked to operate in an image volume comprising segmentations of MRI images reflecting the open space (CSF) and the obstacles. Furthermore, the number of actuation nodes is calculated, *i.e.*, nodes in the path where the robot needs to change its direction, to assess the efficiency of the path. To evaluate the proposed algorithm, a detailed digital model of the SSS was utilized [18]. The 3D model was transformed into a 3D stack of slices depicting an ideal SSS segmented from MRI scans.

The remainder of this paper is organized as follows: Section II presents the soft-robot model used in the context of this work, Section III presents the evaluation process of the proposed algorithm and the path-planning results, and Section IV discusses the conclusions of this work.

## II. METHODOLOGY

This section describes the specification of the robot and its navigation principles, *i.e.*, its design, movement requirements and navigation configuration space. Furthermore, the employment of additional cost functions in the context of A* is analyzed. These cost functions aim at improving the path estimation process in order to satisfy the movement requirements of the robot.

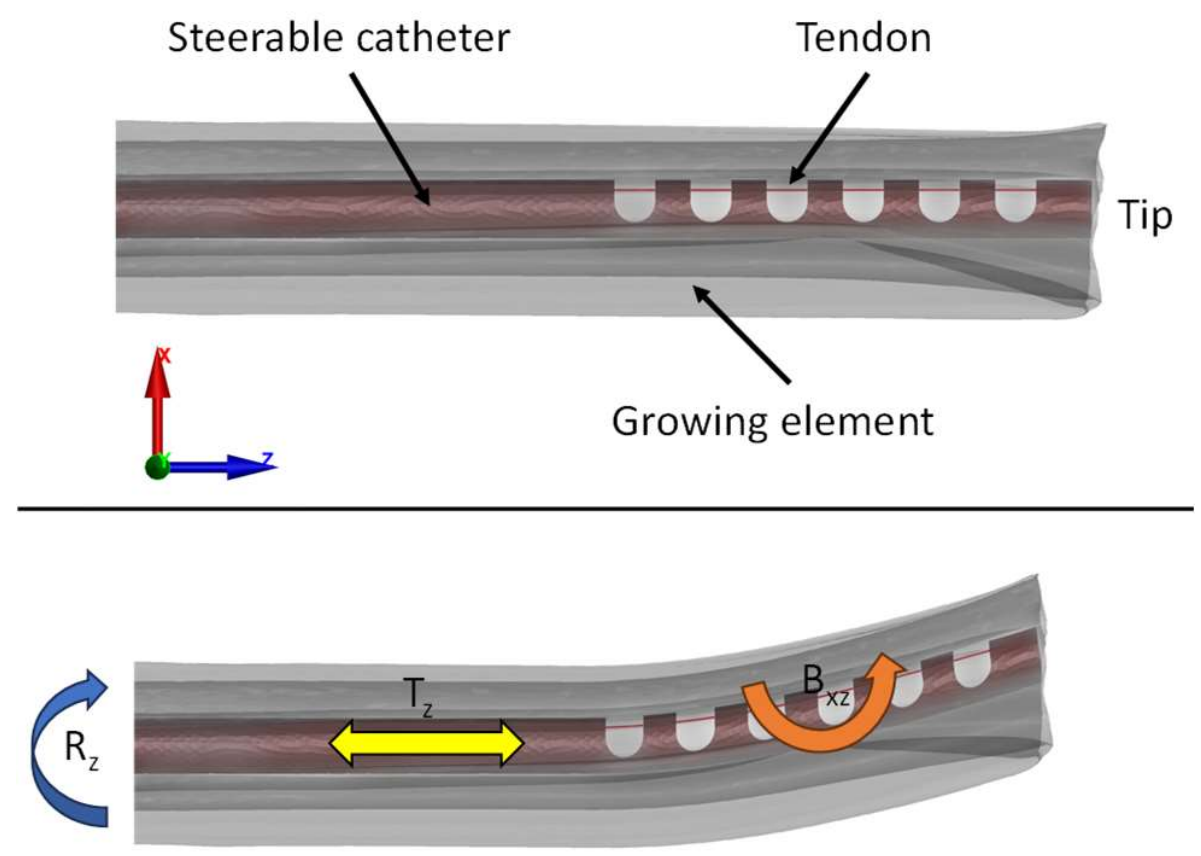

Fig. 1. Illustration of the soft robot composed of the GE and steerable catheter with tendon. When tension is applied to the tendon (bottom image), the front part of the catheter bends. The $R_z$, $T_z$, and $B_{xz}$ arrows demonstrate the possible motions of the catheter, *i.e.*, rotation around the Z-*axis*, translation along the Z-*axis*, and bending over the X-Z plane, respectively.

### A. Soft-Robot Growing & Navigation Principles

Growing robots can navigate confined areas through growth rather than insertion; thus, the shear forces exerted on the surrounding environment are significantly limited. The GE contains a saline solution as the pressurization medium. When the GE is pressurized, its eversion becomes possible with the simultaneous translation of the active channel (attachment point of the GE). The GE features a hollow core that can be utilized as an instrument channel or guide. There, a steering catheter made of a flexible tube with patterned notches on the side of the distal part can be inserted. A tendon allows the distal part of the catheter to bend in a specific direction when it is pulled. By design, the patterned catheter can only bend in one direction, *i.e.*, multi-direction navigation can be achieved through a combination of translation, longitudinal rotation/roll, and tendon-driven bending of the catheter. A schematic illustration of the soft robot structure (distal part) and its motion characteristics are illustrated in Fig. 1. A more detailed description and visualization of the whole robot is provided in [19].

In this work, it is assumed that only the tip of the robot is deployed in the SSS and traverses its length, since the navigation plan is implemented based on the "follow-the-leader" principle, *i.e.*, the path determined by the motion of the tip is the path along which the rest of the catheter moves. Therefore, the rest of the robot body is not considered in this work. The tip of the robot is modeled using a disk and an annulus denoting the catheter and the soft GE, respectively (Fig. 2a):

$$(c_x - x)^2 + (c_y - y)^2 \leq r_c^2 \quad (1)$$

$$r_c^2 \leq (c_x - x)^2 + (c_y - y)^2 \leq r_g^2 \quad (2)$$

where $r_c$ and $r_g$ indicate the radii of the outer circles of the catheter (Eq. (1)) and the GE (Eq. (2)), respectively, the point $(c_x, c_y)$ denotes the coordinates of the co-center, and $(x, y)$ is an arbitrary point on the 2D plane.

Figure 2b illustrates a segmented cross-section of the SSS (white region) volume in which the soft robot is navigated. The black regions indicate the SSS walls (dura and pia mater) as well as nerves and vessels, which are candidate anchor points for the GE. Anchor points are selected along the estimated path where the GE is in contact with surrounding objects marked as possible obstacles. These points occur when contact is established simultaneously at opposite sides of the GE tube. However, if the ratio of the intersection area (orange areas in Fig. 2b) to the total area (blue and orange areas in Fig. 2b) exceeds a specified limit, the path is classified as non-traversable.

### *B. Extended A* Algorithm for Image-based Path-Planning*

To implement the path-planning algorithm in accordance with the scenario described in the previous paragraph, the A* algorithm is extended as follows. The usual implementation of A* evaluates if a node $n = (x, y, z)$, $x, y, z \in Z^+$ should be included in the path using the function $f(n) = g(n) + h(n)$. The cost function $g(\cdot)$ and the heuristic function $h(\cdot)$ represent the cost from a node $n$ to the starting and ending points, respectively. These cost functions usually estimate the Euclidean distance from a node $n$ to the start node $n_s$ and the target node $n_g$:

$$g(n) = d(n_s, n) \tag{3}$$

$$h(n) = d(n_g, n) \tag{4}$$

where $d(\cdot)$ is the function estimating the Euclidean distance between two points. Due to its heuristic nature, the A* algorithm can generate paths that run near obstacles [20]. In our application, this behavior is acceptable; however, it does not guarantee the establishment of anchor points. To tackle this issue, an additional parametric cost function $c_a(\cdot)$ is introduced to award nodes that can provide anchor points for the robot. This function is defined as:

$$c_a(n, k(n), z) = e^{\frac{k(n)}{z}} - 1 \tag{5}$$

where $k(n)$ is an adaptive parameter that changes according to $n$:

$$k(n) = \begin{cases} k(n) = 0, if\ n \text{ is anchored} \\ k(n) = k(n) + 1, if\ n \text{ is not anchored} \end{cases} \tag{6}$$

and $z \in \mathbb{R}^+$ configures the steepness of the function $c_a(\cdot)$. Furthermore, an additional cost function is considered to further drive the robot near obstacles. The cost function $c_b(\cdot)$ estimates the distance of a node $n$ to the closest point classified as an obstacle; hence, it penalizes nodes that are far from obstacles.

Considering the above, the evaluation function of the proposed path-planning algorithm can be formed as:

$$f(n) = g(n) + h(n) + c_a(n; k(n), z) + c_b \tag{8}$$

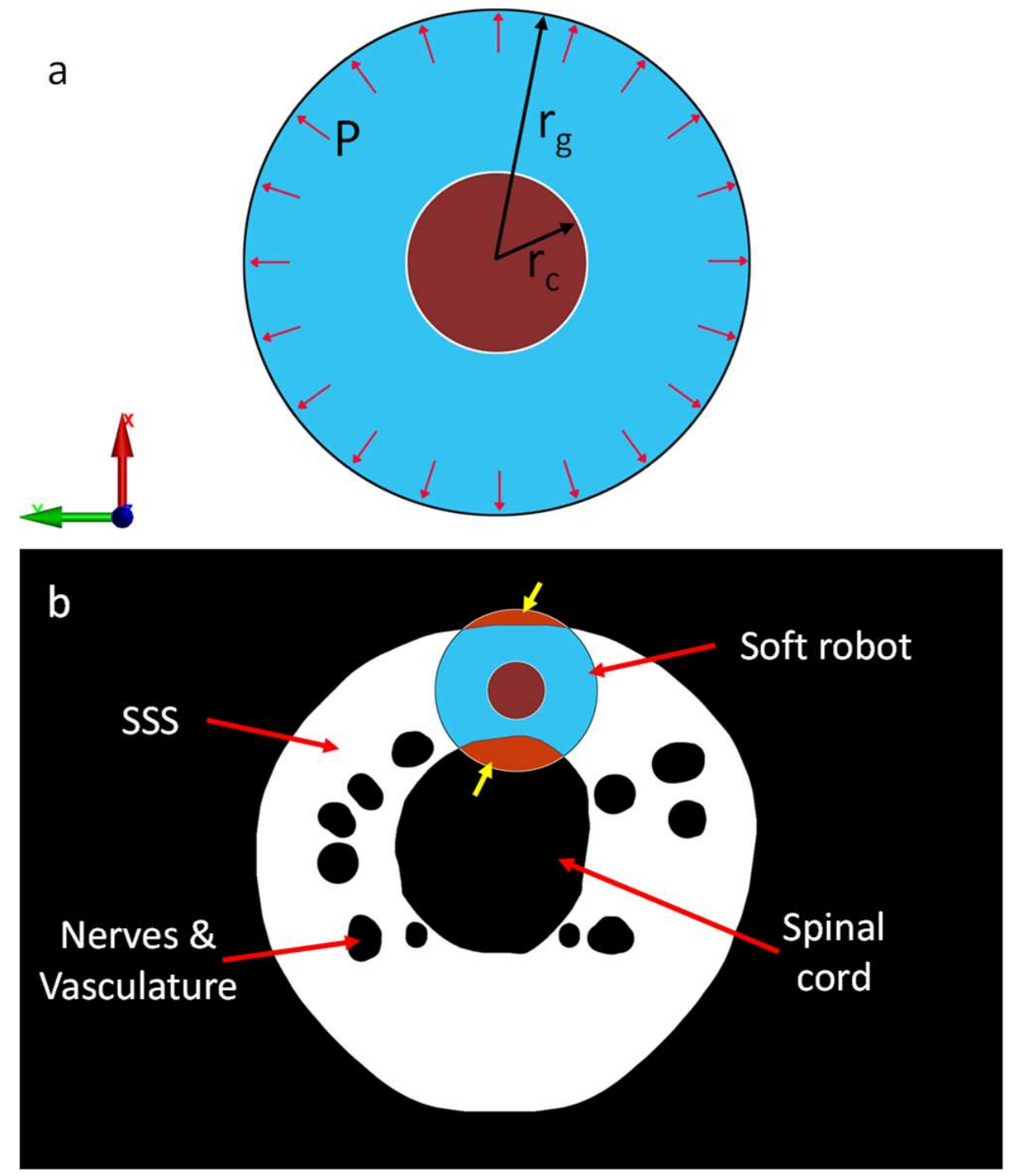

Fig. 2 (a) Schematic illustration of a cross-section of the soft robot, where P denotes the pressure applied by the contained fluid (saline solution), $r_g$ is the GE radius, and $r_c$ is the catheter radius; (b) Superposition of (a) on a segmented MRI slice of the SSS. The areas indicated by the yellow arrows represent the overlapping regions used as indicator for the safe anchoring of the robot.

### *C. Navigation Configuration Space*

As described in Section II-A, the soft robot is navigated based on the information that can be derived regarding the surrounding volume, including the possible obstacles and the free space of the SSS. This volume is obtained from a stack of segmented images, which is denoted as $\boldsymbol{S} \in \{0, 1\}^{H \times W \times N}$, where 1 and 0 denote the presence or absence of anatomical structures, *i.e.*, probable obstacles or the free space, respectively. The notations $H$ and $W$ correspond to the height and width of each segmentation map, whereas $N$ indicates the total number of segmentation maps in the stack.

The configuration space $C$ comprises all the points $q_i = (x_i, y_i, z_i)$, $i = 1, 2, 3, \ldots, H \times W \times N$, $x, y, z \in Z^+$ indicating a unique position in $\boldsymbol{S}$. The origin of the coordinate system used is the top left corner of the first slice of the image stack, $x_i$ and $y_i$ are coordinates indicating a point along the width and height of a slice, whereas $z_i$ is a coordinate that indicates the position of a slice along the depth of the stack. Moreover, $C$ can be divided into two subspaces, $C_{free}$ and $C_{obs}$, each including $q$ that belong to the free ($\boldsymbol{S}(q) = 0$) and obstacle ($\boldsymbol{S}(q) = 1$) spaces, respectively. However, as it can be derived from Eqs. (1) and (2), the catheter and GE of the soft robot do not share the same free and obstacle spaces. Considering Eq. (1) and the model of the robot, the free space of the catheter is defined as:

$$C_{free}^c = \{q = (x, y, z) | \forall q \in C, S(q) = 0, \|q - p\|_2 > r_c\} \tag{9}$$

where $p$ denotes these $q \in C$ that $\boldsymbol{S}(q) = 1$. Accordingly, the obstacle space of the catheter is:

$$C_{obs}^{c} = \{q = (x, y, z) | \forall q \in C, S(q) = 1, (c_x - x)^2 + (c_y - y)^2 \leq r_c^2\} \quad (10)$$

Nonetheless, the free and obstacle spaces of the GE are parameterized by its design and other factors, *e.g.*, pressure $P$. A point $q$ designates a possible path node for the GE when the ratio of the intersection area, which includes points denoting obstacle locations, to the total area is below a value $v$:

$$C_{free}^{g} = \{q = (x, y, z) | \forall q \in C, \frac{\sum_{q \in area(q)} S(q)}{|area(q)|} \leq v\} \quad (11)$$

$$C_{obs}^{g} = \{q = (x, y, z) | \forall q \in C, \frac{\sum_{q \in area(q)} S(q)}{|area(q)|} \geq v\} \quad (12)$$

where $area(q)$ represents all the points that are inside the GE annulus, as defined in Eq. (2), given that $q$ is the center and $|area(q)|$ denotes the total number of points in the GE. Regarding the configuration space of anchor points, they are denoted as the nodes where two opposite sides of the GE come into contact with obstacle regions (Fig. 2). The opposite regions are denoted by opposite arcs of the outer circle of the GE annulus. Hence, the outer circle of the annulus is divided into four arcs and the membership of a point $q$ to an arc is defined as follows:

$$\boldsymbol{right}: \boldsymbol{if}\ tan^{-1}\left(\frac{y - c_y}{x - c_y}\right) \in [0, \frac{\pi}{4}) \cup [\frac{7\pi}{4}, 2\pi] \quad (13)$$

$$\boldsymbol{left}: \boldsymbol{if}\ tan^{-1}\left(\frac{y - c_y}{x - c_y}\right) \in [\frac{3\pi}{4}, \frac{5\pi}{4}) \quad (14)$$

$$\boldsymbol{up}: \boldsymbol{if}\ tan^{-1}\left(\frac{y - c_y}{x - c_y}\right) \in [\frac{\pi}{4}, \frac{3\pi}{4}) \quad (15)$$

$$\boldsymbol{down}: \boldsymbol{if}\ tan^{-1}\left(\frac{y - c_y}{x - c_y}\right) \in [\frac{5\pi}{4}, \frac{7\pi}{4}) \quad (16)$$

Hence, if the GE at a node $n$ is in contact with multiple points $q \in C_{obs}^{g}$ that are either *left* and *right* or *up* and *down*, this node is classified as an anchor point.

### D. Estimation of Actuation Nodes

To assess the efficiency and capacity of the robot to navigate along the estimated path, the nodes that require the actuation of the catheter are estimated. The detection of actuation nodes is performed by estimating the angle $a_{act}$ of the current node $n_i$ with respect to the previous and consecutive nodes, $n_{i-1}$ and $n_{i+1}$, respectively, where $i = 1, 2, 3 \ldots M$ indicates the position of a node in a path and $M$ denotes the total number of nodes. If $a_{act}$ is non-zero, then, that node is classified as an actuation node.

Given three consecutive nodes in a path, *i.e.*, $n_{i-1}$, $n_i$, and $n_{i+1}$, to determine if node $n_i$ is an actuation node, the value of $a_{act}$ is estimated as follows:

$$a_{act} = \cos^{-1}\left(\frac{d(n_{i-1,i})^2 + d(n_{i,i+1})^2 - d(n_{i-1,i+1})^2}{2 \cdot d(n_{i-1,i})^2 \cdot d(n_{i,i+1})^2}\right) \quad (17)$$

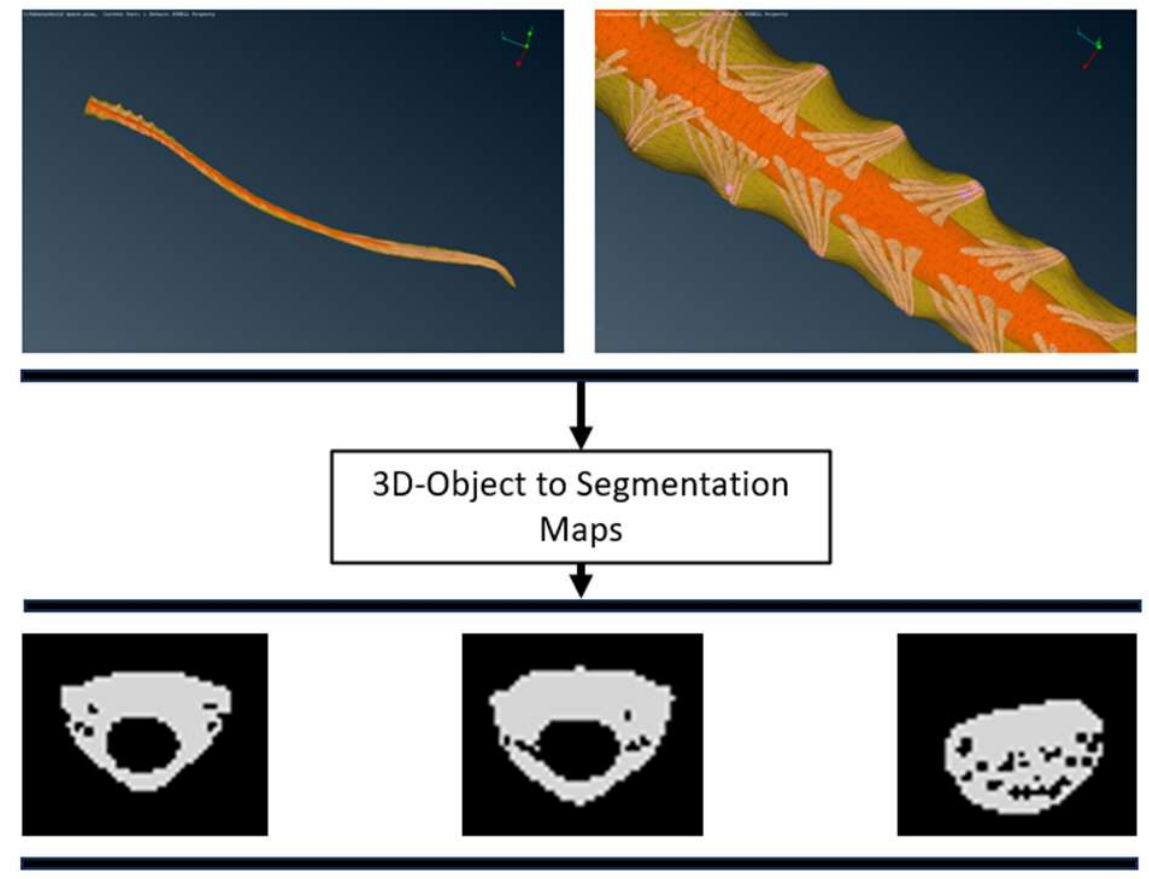

Fig. 3. Visual example of the transformation of the 3D SSS object into an image stack of segmentation maps.

TABLE I. QUANTITATIVE COMPARISON OF THE PATH-PLANNING

| Methods | Metrics | | |
|---|---|---|---|
| | *APAP (%)*↑ | *ADT (mm)*↓ | *APAN (%)*↓ |
| $A^*$ | 14.37 ± 12.6 | 1.09 ± 0.17 | 29.26 ± 5.36 |
| $A^* + c_a$ | 19.41 ± 14.72 | 1.10 ± 0.18 | 32.87 ± 5.18 |
| $A^* + c_b$ | 15.19 ± 13.08 | 1.31 ± 0.19 | 37.68 ± 4.53 |
| $A^* + c_a + c_b$ | 19.11 ± 15.31 | 1.35 ± 0.19 | 38.83 ± 4.54 |

↑ and ↓ indicate that higher and lower values are better, respectively.

## III. RESULTS

### A. Experimental Setup

First, a digital model of the SSS was obtained from [18]. Subsequently, the 3D model was imported into the 3D Slicer software where it was transformed into a stack of images possessing similar characteristics with segmentation maps derived from the processing of MRI scans. In this way, the time required for validating the proposed algorithm on actual data was significantly shortened.

Each segmentation was extracted in a way to maintain the calibration parameters of the original model, *i.e.*, each slice had an in-plane resolution of 0.547×0.547 $mm^2$ and a slice thickness of 1 mm. These parameters resulted in the generation of 574 segmentation maps with a resolution of 223×128. However, the utilized model included segments in which our robot is not intended to operate. Thus, these segments were discarded, resulting in 413 segmentation maps that were used for the evaluation process.

The radii of the catheter and GE were set to 0.5 mm and 1 mm, respectively. The GE safe passage $v$ was set to 80%. The steepness factor $z$ of $k(n)$ was set to 5.

### B. Quantitative Results

To quantify the efficiency of the estimated paths, three metrics were incorporated: the average percentage of anchor points (APAP) in the path, the average distance travelled (ADT) between consecutive nodes, and the average percentage of actuation nodes (APAN). Higher APAP values ensure a more stable maneuvering of the robot, whereas lower ADT and APAN values indicate paths of lower complexity. Furthermore, for comparison purposes, the proposed methodology was compared to the original implementation of the A* algorithm, *i.e.*, leveraging only the $g(\cdot)$ and $h(\cdot)$ functions, as well as the A* algorithm with only the $c_a$ and $c_b$ cost functions.

To experimentally evaluate the proposed methodology on the SSS model, a total of 200 runs were performed. At each run, two points were randomly selected as the start and end points from the first and last segmentation map of the stack, respectively. These points were bound to belong to the free space of the catheter and GE. Table 1 summarizes the results obtained from the evaluation process. As it can be observed, both the A* variants, leveraging all the cost functions and using individually the $c_a$ achieved the highest percentage of anchor points in a path. However, solely the use of $c_a$ led to a lower percentage of actuation nodes, indicating improved navigation efficiency. Even though the simple A* appears to perform better with respect to ADT and APAN, its APAP value is lower; thus, there should be a balance between the three metrics to ensure the feasibility of the robot navigation through robust anchoring and paths of reasonable complexity.

### C. Qualitative Results

To further verify that the estimated path runs properly through the SSS, avoiding collision of the catheter with obstacles while exploiting them to ensure that the GE is securely anchored to the surrounding structures, the obtained path (points) was introduced into the Siemens NX software (v11) together with the digital model of the SSS. Subsequently, the path points were transformed into a spline, which was then used to generate a tube with a diameter of 2 mm, *i.e.*, equal to the outer diameter of the GE.

Figure 4 presents the superposition result of the obtained path visualized as the GE tube and the SSS model. It can be observed that the soft robot could run safely through the SSS, avoiding all obstacles. In particular, it can be seen in Fig. 4a that the soft robot avoided the nerve roots when passing from the anterior to the posterior part of the SSS. This is also illustrated in Fig. 4c, where the dura mater has been removed for better visualization. Furthermore, Fig. 4b demonstrates the effectiveness of the proposed algorithm, since the intersection lines (blue lines) reveal that the GE could be stably anchored on the SSS walls, facilitating the steering of the catheter and ensuring that the "follow-the-leader" principle is upheld. From the visualization of the path, it can be observed that the direction changes of the robot appear rather abrupt; however, these can be easily corrected via post-processing smoothing procedures without affecting the properties of the path, *e.g.*, anchoring points.

In addition, the path obtained by the proposed algorithm was compared to that obtained by the traditional A* algorithm under the same conditions. Figure 5 illustrates the comparison result at a cross-section of the SSS model, where it can be observed that the proposed algorithm managed to successfully anchor the soft robot at opposite sides of the GE tube. On the other hand, the traditional A* algorithm failed to satisfy this requirement, introducing locations where the soft robot would be able to move freely in the SSS, compromising the steering capability of the catheter, and as a result, the successful navigation along the generated path.

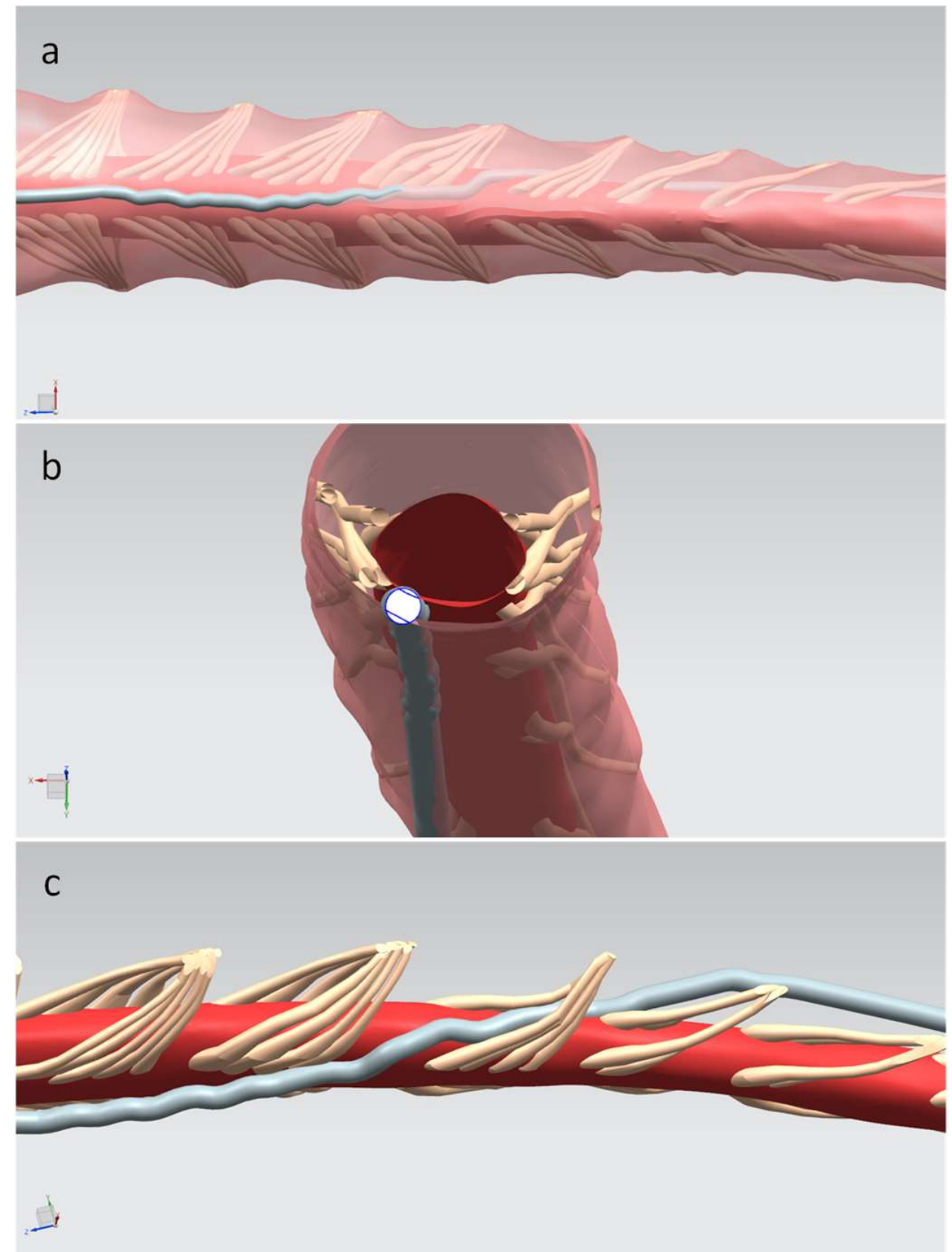

Fig. 4 Visualization of an estimated path (GE tube) in the SSS digital model. Structures in pink color depict the dura mater, white the nerve roots, and red the spinal cord. (a) Posterior view of the GE tube running through the SSS; (b) Cross-section of the model, depicting the intersection of the GE tube with the surrounding structures (blue lines); (c) Side view depicting the soft robot passing through the nerves

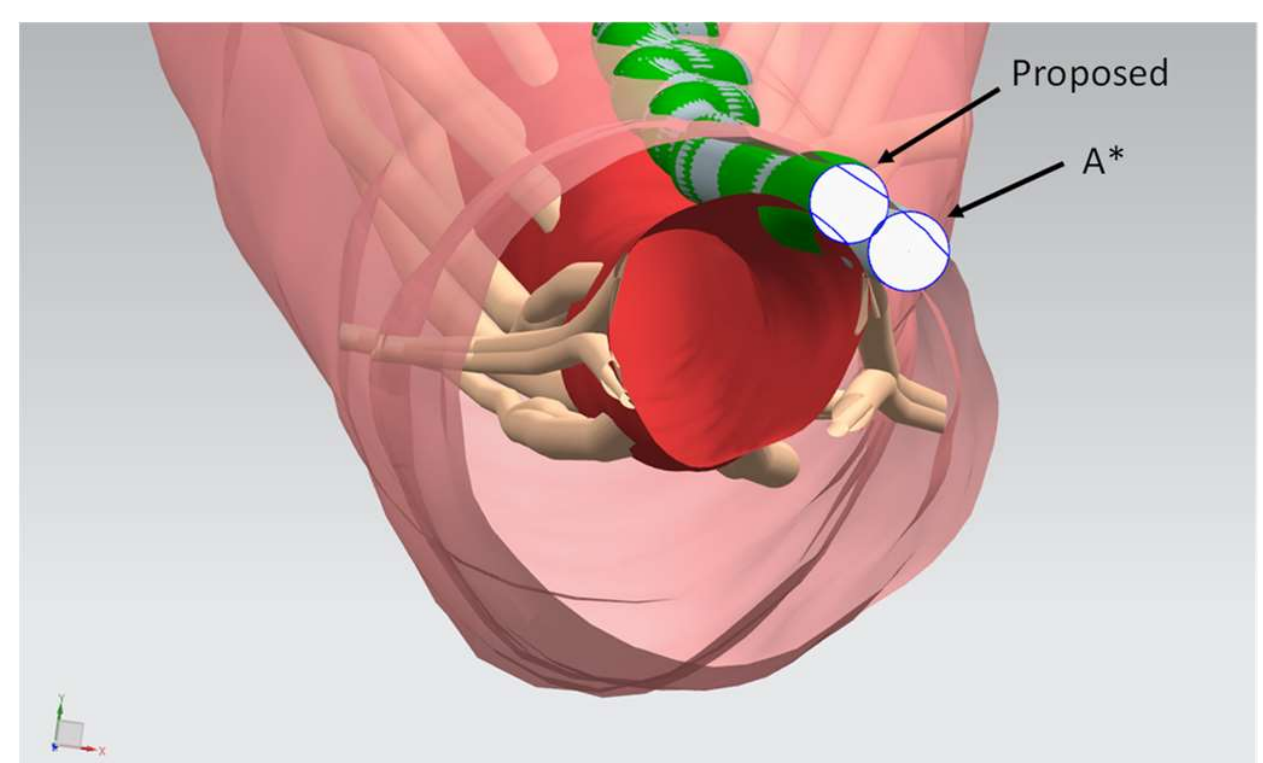

Fig. 5 Cross-sectional view comparing the paths (GE tubes) estimated by the proposed and A* algorithms. The blue lines indicate the intersections with the surrounding structures that ensure safe anchoring of the GE.

## IV. Conclusions

This study presented an image-based path-planning algorithm specifically designed for guiding soft-growing robots within the complex environment of the SSS. By extending the capabilities of the A* algorithm, the incorporation of additional cost functions was investigated to estimate safe pathways towards targeted locations and determine essential anchor points for the effective steering and maneuverability of the soft-growing robot. The proposed algorithm was evaluated based on a highly detailed anatomical model of the SSS, which demonstrated the capacity of the additional cost functions to enable the navigation of the robot through delicate tissue structures, while fulfilling the movement requirements of the robot and maintaining the path efficiency. These findings underscore the potential of the incorporation of the additional cost functions in advancing soft robotic technologies for minimally invasive medical procedures, offering a promising avenue for enhanced targeted drug delivery and operations.

Future work will focus on further refining the proposed algorithm by evaluating its applicability utilizing SSS phantoms, real world data (*i.e.*, segmentations derived from MRI scans of different individuals), and by conducting finite element analysis simulations of the robot navigating through the estimated path. In addition, the kinematics of the robot that were not considered in this study, will be incorporated into the path-planning algorithm in subsequent research steps. Further path-planning approaches, *e.g.*, reinforcement-learning and optimization-based solutions, will also be in the scope of future work.